\documentclass[journal]{IEEEtran}
\usepackage[sorting=none, style=ieee]{biblatex}
\usepackage{amsmath,amsfonts}
\newcommand{\sign}{\text{sgn}}
\usepackage{algorithmic}
\usepackage{algorithm}
\usepackage{array}
\usepackage{textcomp}
\usepackage{stfloats}
\usepackage{url}
\usepackage{bm}
\usepackage{siunitx}
\usepackage{svg}
\usepackage{verbatim}
\usepackage{graphicx}
\usepackage{subcaption}
\usepackage{printlen}

\usepackage{siunitx}
\begin{document}
%\printlength\textwidth

\title{Interaction-aware Decision-making for Automated Vehicles using Social Value Orientation}

\author{Luca Crosato, Hubert P. H. Shum, Edmond S. L. Ho, and Chongfeng Wei
        % <-this % stops a space
\thanks{Luca Crosato is with both Queen's University Belfast and Northumbria University (email: l.crosato@qub.ac.uk, luca.crosato@northumbria.ac.uk).}
\thanks{Hubert P. H. Shum is with the Department of Computer Science, Durham University, UK (email: hubert.shum@durham.ac.uk).}
\thanks{Edmond S. L. Ho is with the Department of Computer and Information Sciences, Northumbria University, UK (email: e.ho@northumbria.ac.uk).}
\thanks{Chongfeng Wei is with the School of Mechanical and Aerospace Engineering at Queen's University Belfast, UK (email: c.wei@qub.ac.uk).}
\thanks{Corresponding author:Dr. Chongfeng Wei}\vspace{-0.5cm}}

\markboth{}%
{Crosato \MakeLowercase{\textit{et al.}}: Interaction-aware Decision-making for Automated Vehicles using Social Value Orientation}

\IEEEpubid{0000--0000/00\$00.00~\copyright~2022 IEEE}

\maketitle

\begin{abstract}
Motion control algorithms in the presence of pedestrians are critical for the development of safe and reliable Autonomous Vehicles (AVs).
Traditional motion control algorithms rely on manually designed decision-making policies which neglect the mutual interactions between AVs and pedestrians. On the other hand, recent advances in Deep Reinforcement Learning allow for the automatic learning of policies without manual designs.
To tackle the problem of decision-making in the presence of pedestrians, the authors introduce a framework based on Social Value Orientation and Deep Reinforcement Learning (DRL) that is capable of generating decision-making policies with different driving styles. 
The policy is trained using state-of-the-art DRL algorithms in a simulated environment. A novel computationally-efficient pedestrian model that is suitable for DRL training is also introduced. We perform experiments to validate our framework and we conduct a comparative analysis of the policies obtained with two different model-free Deep Reinforcement Learning Algorithms. Simulations results show how the developed model exhibits natural driving behaviours, such as short-stopping, to facilitate the pedestrian's crossing.
\end{abstract}

\vspace{-0.1cm}
\section{Introduction}
\IEEEPARstart{A}{utonomous} driving is an emerging technology with the potential to drastically impact our society. The most notable benefit that Autonomous Vehicles (AV) can bring about in our everyday lives is the vast reduction of vehicular accident risks \cite{brenner2018overview}. Over the past few years, AV research has attracted the attention of both industry and academics with a rapid uptake of Advanced Driver-Assistance Systems (ADAS). Despite the recent advancements, the number of recorded road deaths worldwide remains extremely high. In 2018, the World Health Organisation reported 1.35 million deaths arising from traffic accidents \cite{world2018global}, which provides a clear motivation for the development of Autonomous Vehicles that ensure the safety of all road users, including pedestrians.

One of the major challenges in autonomous driving is achieving \textit{collision-free navigation} in cluttered and interactive environments in the presence of pedestrians. 
This challenge involves finding an optimal path between the vehicle's current and target location that minimises journey time and is guaranteed to be collision-free, while satisfying constraints imposed by the AV mechanics \cite{paden2016survey}.
While traditional motion control algorithms offer a solution to this problem, they typically suffer from two main drawbacks. Firstly, they can be overly cautious when interacting with pedestrians compared to an average human driver \cite{seth2019traffic}, \cite{trautman2010unfreezing}. This behaviour results in an unpredictable driving style, potentially leading to accidents. Secondly, they struggle to adapt to unseen situations, which represents an obvious problem given the number of possible road scenarios is countless.

\begin{figure}[tp]
\centering
    \includegraphics[width=0.38\textwidth]{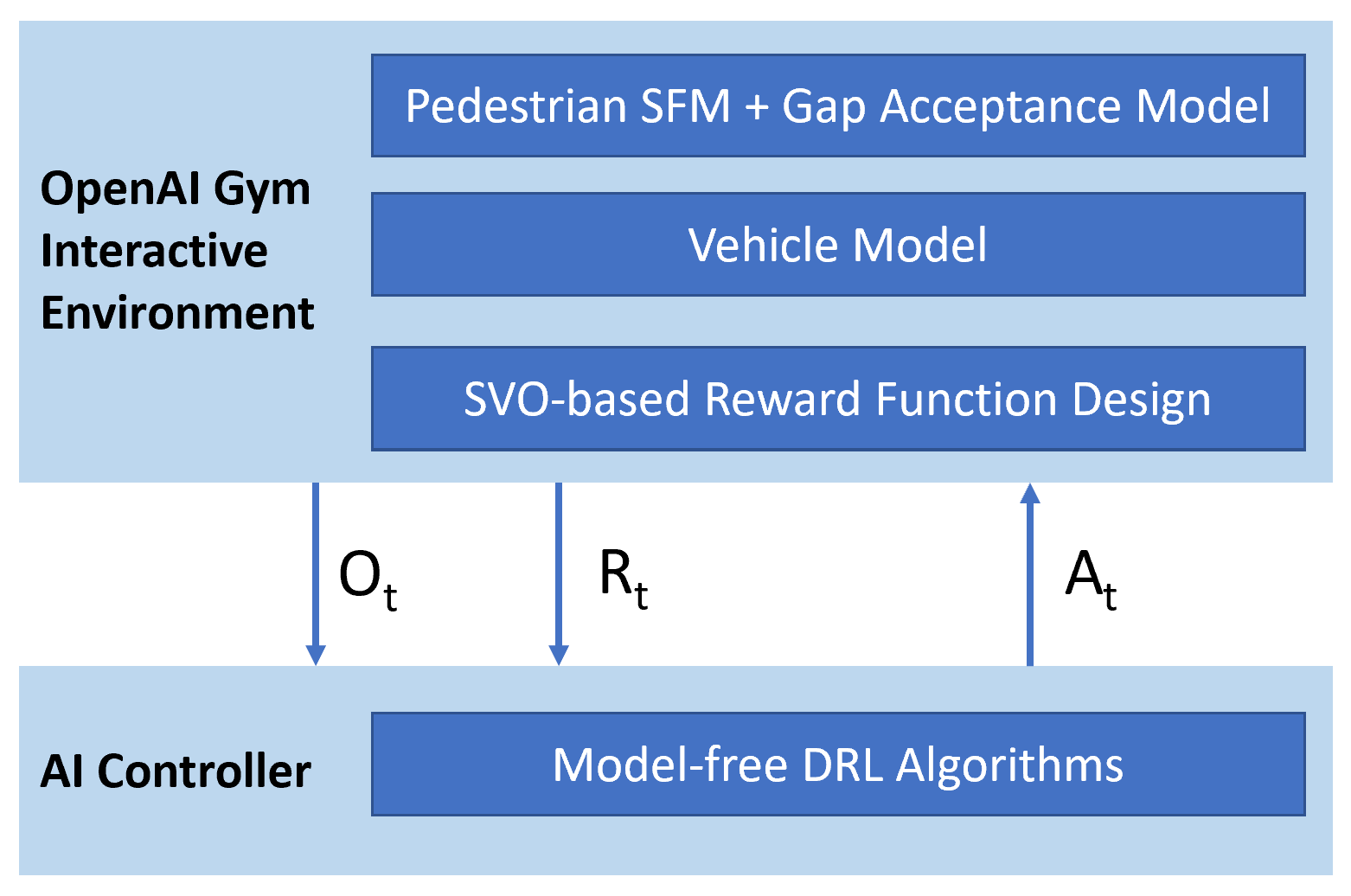}
    \caption{Technical framework used. $O_t$, $R_t$, $A_t$ represents reinforcement learning observations, reward, and actions respectively.}
    \label{fig:technical-framework}
    \vspace{-0.7cm}
\end{figure}
In this paper, we propose to solve the above issues by adopting a learning-based approach and by utilising the concept of Social Value Orientation (SVO) from Social Psychology \cite{murphy2011measuring, van1997development, mcclintock1989social} into the AV motion controller design. SVO is a value that quantifies how much a person values the welfare of the others compared to their own.
We employ Deep Reinforcement Learning (DRL), a subfield of machine learning that combines Reinforcement Learning and Deep Learning. We are witnessing an increasing number of publications seeking to utilise DRL to solve Autonomous Driving problems \cite{li2021safe, ma2021reinforcement}. In DRL, a motion controller is synthesised through a trial-and-error process in a safe simulated environment without the need to manually handcraft an AV decision-making policy, making maintenance and development simple. For a more comprehensive review on the topic see \cite{aradi2020survey}.
%\color{blue}
Existing DRL methods only focus on the ego-vehicle own goals, neglecting the fact that its actions may negatively impact surrounding vehicles. In particular, the reward function design only takes the ego-vehicle goals into account. Our novelty is the shaping of the DRL reward function with SVO to take the surrounding road users' comfort into account. 
\IEEEpubidadjcol

We argue that studying the mutual interactions between an AV and a single pedestrian is the first essential step in the development of Vulnerable Road User's friendly policies for Autonomous Vehicles, therefore we consider a typical straight road scenario with one pedestrian and train a set of DRL agents with different SVO values.
Our proposed framework is shown in Fig. \ref{fig:technical-framework}.
First, we develop a novel interactive pedestrian model that combines the concepts of situational awareness \cite{markkula2018models} and Social-Force \cite{helbing1995social} to determine the pedestrian trajectory under the vehicle influence. The vehicle motion affects the pedestrian decisions by indirectly altering the available time-gap to complete crossing and the social forces acting on the pedestrian. In turn, pedestrian motion serves as a cue for the ego-vehicle controller, thereby mutually influencing each other.  We evaluate our pedestrian model using a set of typical road scenarios and by comparing pedestrian motion statistics with real world data and a state-of-the-art pedestrian model. 
Then, agents trained with model-free DRL algorithms learn the interaction patterns with the pedestrian and exploit them to indirectly affect pedestrian motion. For instance, the vehicle learns the effect that its own acceleration on pedestrian's decisions, thereby hindering or favouring the pedestrian crossing. We demonstrate how our reward choice produces controllers that naturally exhibit human-like behaviour, with a plethora of different driving styles, ranging across a spectrum from aggressive to pro-social according to the choice of the SVO value. 
We conduct a set of qualitative and quantitative experiments aimed at evaluating the effect of SVO addition, and model performances under both nominal and high-risk scenarios.  
\color{black}
%Since we are training our model in a simulated environment and we are looking to deploy it in the real world in the future, it is important that the simulated pedestrian behaviour be as similar as possible to that of a real one. 

%The preliminary result of this research has been published as a conference paper \cite{luca21humancentric}. Since this initial report, we have made a number of improvements. First and foremost, we have introduced a novel pedestrian model that demonstrates more realistic pedestrian behaviour. We have retrained our RL vehicle controller and validated that it is effective in interacting with pedestrians with increased complexity. Finally, we have conducted the same experiments as in \cite{luca21humancentric} to evaluate our controllers, with an added comparative analysis of two model-free DRL algorithms, specifically the Soft Actor Critic (SAC) and Proximal Policy Optimisation (PPO) algorithms. 
Coincisely, this paper presents the following contributions: 
\begin{enumerate}
%first, 
%\item We propose a new framework for training the agent AV with state-of-the-art RL algorithms to solve the pedestrian collision avoidance problem in structured scenarios;
\item We demonstrate how the introduction of SVO into the DRL Reward Function design influences the ego-vehicle strategies, achieving behaviours that range from egoistic to pro-social, without affecting pedestrian safety;
\item We introduce a novel pedestrian simulation model that combines gap-acceptance methods with Social Force Models to model the pedestrian crossing behaviour;
\item We validate that our RL model is capable of handling the added complexity introduced by our more realistic pedestrian model that actively reasons about the AV's actions and conduct a comparative analysis of two model-free DRL algorithms applied to our problem.
\end{enumerate}

\vspace{-0.1cm}
\section{Related Work}
This section is organised as follows. First, we will review state-of-the-art motion-planning and decision-making techniques. As learning-based techniques are more relevant to this work, we will mainly focus on them (see \cite{schwarting2018planning} for a more comprehensive review on motion-planning for AVs). Secondly, we will review pedestrian models in autonomous driving that can be used in AV-pedestrian interaction simulations. Unlike behavioural cloning methods \cite{ly2020learning}, DRL allows an autonomous vehicle to exceed human-level performances by optimisation of a reward function. One of the first applications of DRL to the field of autonomous driving can be found in \cite{vitelli2016carma}. In this work, RL was used to train an AV in a recing environment. RL has also been used to learn an automated lane change policy in \cite{min2019deep, hoel2019combining, wang2019lane} and to solve complex urban navigation problems. Chen \textit{et al.} \cite{chen2019model} employed DRL to a roundabout scenario with multiple vehicles. Their method is trained and evaluated in the open-source driving simulator CARLA \cite{dosovitskiy2017carla}. Sallab \textit{et al.} \cite{sallab2017deep} have used Recurrent Neural Networks in the Reinforcement Learning framework to account for Partially Observable scenarios and integrated attention models in the framework, making use of attention networks to reduce computational complexity. 

\begin{figure}[t]
\centering
    \includegraphics[width=1.0\linewidth]{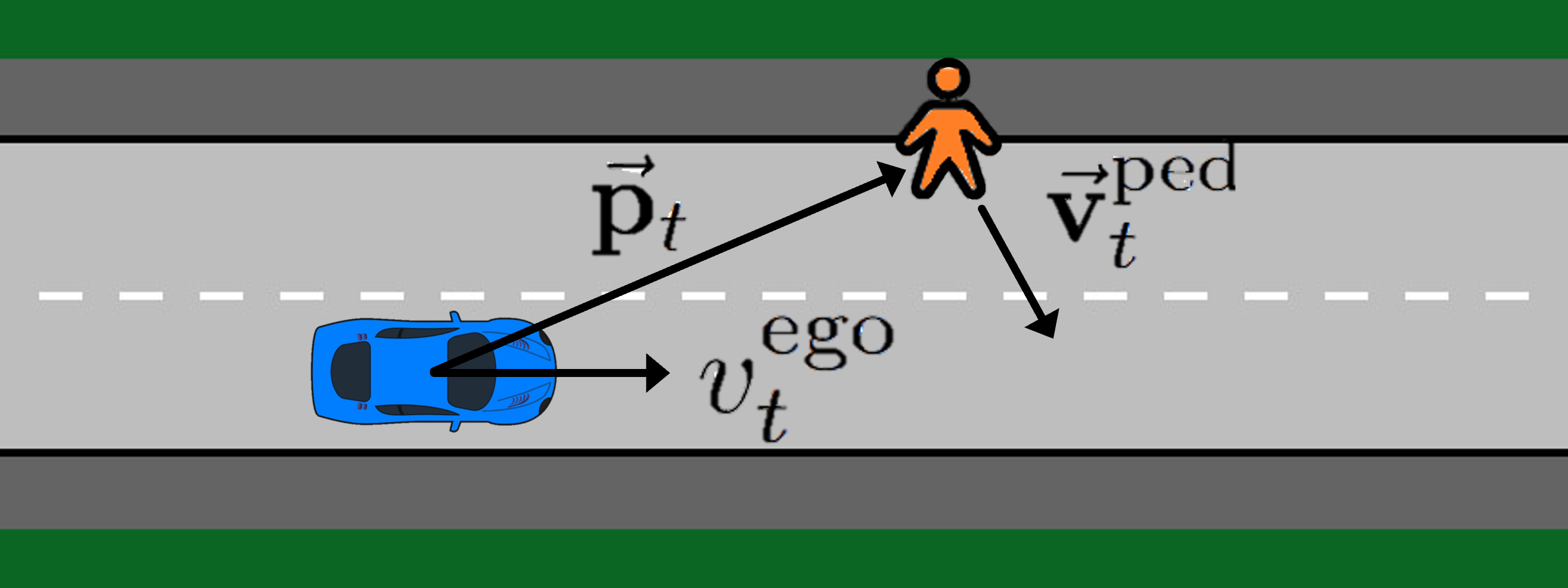}
    \caption{Scenario illustration.}
  \label{fig:scene}
    \vspace{-.6 cm}
\end{figure}

Applications of DRL in human-AV interactive scenarios is also an emerging field of research, with applications in mobile robot navigation amongst crowds. Gao et al. \cite{gao2017intention} trained a local neural network motion controller with DRL to avoid moving obstacles such as pedestrians and combined it with a global path planner. In \cite{everett2021collision}, DRL is used to develop an algorithm that learns a collision avoidance policy among a variety of heterogeneous, non-communicating, dynamic agents. The authors demonstrate how a mobile robot trained with this policy is capable of travelling at walking speed in a crowd.

The applications of DRL to the problem of collision avoidance in structured scenarios with crossing pedestrians is more limited. Structured scenarios arises when the AV and the pedestrians navigate in separate areas for most of the time with temporary areas of conflicts. Deshpande \textit{et al.} \cite{deshpande2019deep} trained a Deep Q-Network vehicle agent at a typical intersection crossing scenario. The pedestrian information is represented in a grid-based approach as a state space input to the learning agent. 
In \cite{li2020deep} the authors trained a DQN policy to avoid pedestrians and used it to develop a driving assistance system to assist human drivers in the event where their behaviour is dangerous for the pedestrian. One of the main limitations of DQNs is the discretisation of the action space, which we overcome in our work with continuous action policies.

Social Psychology has shown each individual has personal preferences on how to value their utility in relation to that of the others. SVO can predict negotiation strategies and cooperation motives in games involving multiple individuals \cite{murphy2011measuring, mcclintock1989social, van1997development, ackermann2019explaining}. 
SVO has been used in conjunction with Inverse Reinforcement Learning \cite{schwarting2019social} to estimate other surrounding drivers' behaviour and improve trajectory prediction, while Sun \textit{et al.} \cite{sun2018courteous} developed a courteous AV model based on Inverse Reinforcement Learning for lane merging scenarios. In our work, we present a DRL training framework with Social Value Orientation to model courteous AV behaviours. Instead of using Inverse Reinforcement Learning which would require to collect a big amount of data, we choose to directly use Deep Reinforcement Learning. This also allows us to tailor the reward function to the vehicle-pedestrian interaction scenario.
Since the pedestrian model will be extensively used to train our DRL agent, we need it to possess three main characteristics: firstly, we require it to be computationally efficient to avoid bottlenecks during training; secondly we need it to be realistic; and finally we need the pedestrian to actively reason about the AV's decision, so as to achieve an interactive behaviour that can be learnt and exploited by the AV agent. In this paper we propose a novel pedestrian model that integrates the concept of situational awareness into the Social Force Model framework to achieve an interactive behaviour that explicitly reasons about the AV's actions.

Several works have studied the vehicle-pedestrian interaction in crossing scenarios.
A comprehensive review on pedestrian models in autonomous driving can be found in \cite{camara2020pedestrian1} and \cite{camara2020pedestrian2}. Gap acceptance is a major factor that influences pedestrian decision at intersections. Gap acceptance models have been used to describe the probabilities of pedestrians crossing in a certain gap between vehicles \cite{schroeder2014empirically, sun2003modeling, tian2020creating, tian2022explaining}. These models are used to describe the pedestrian crossing probability but they do not model the trajectories that the pedestrian will follow. 
Markkula \textit{et al.} \cite{markkula2018models} introduced the concept of situational awareness in the pedestrian crossing modelling. In their model, the authors describe pedestrian road crossing decision as the result of a number of perceptual decisions concerning the available gap. A limitation in pure gap-acceptance models is the assumption that once a pedestrian initiates crossing, they will follow a constant speed velocity profile. 

On the other hand, Social Force Models \cite{helbing1995social} of pedestrian behaviour describe collective behaviours by modelling how each individual interacts with other. The idea behind this model is that the influence of surrounding agents on the pedestrian motion can be modelled with forces that measure for the internal
motivations of the individuals to perform certain movements. This model was originally designed for simulating crowd dynamics but has been extended with the effect of vehicles on pedestrians \cite{zeng2014application, yang2021sub}, which makes them suitable for mixed scenarios containing both vehicles and pedestrians. 
Existing works \cite{yang2020multi, ningbo2017simulation} in pedestrian simulation combined social force models with a rule based approach for pedestrian crossing simulation. These models however mainly focus on situations in which pedestrians are in front of the vehicle. In our DRL setting, especially when the vehicle policy is not yet trained, episodes in which the vehicle and the pedestrian are next to each other will be present, which is why we extend the pedestrian model to such scenarios. Secondly, we add a temporal aspect to the decision making process, by including situational awareness in the pedestrian decision making.
\vspace{-0.3cm}

%% Section
\section{Technical Background}
\subsection{Social Force Models}
In a Social Force Based model \cite{helbing1995social}, pedestrians are regarded as point mass particles, with their motion governed by Newton equations of motion:
\begin{equation}
    \frac{d^2 \Vec{\mathbf{r}}}{dt^2} = \frac{d \Vec{\mathbf{v}}}{dt} = \frac{\Vec{\mathbf{F}}_{total}}{m}
\end{equation}
where $\Vec{\mathbf{r}}$ and $\Vec{\mathbf{v}}$ represent the pedestrian position and velocity respectively, and $m$ represents the pedestrian's mass. The total force $\Vec{\mathbf{F}}_{total}$ influences the pedestrian acceleration and can be decomposed further in three terms:
\begin{equation}
    \Vec{\mathbf{F}}_{total} = \Vec{\mathbf{F}}_{nav} + \Vec{\mathbf{F}}_{veh} + \Vec{\mathbf{F}}_{soc}
\end{equation}

The three terms have different effects on the pedestrian motion and shape the pedestrian's trajectory, making them reach their goal position while avoiding obstacles at the same time.
The term $\Vec{\mathbf{F}}_{nav}$ has the overall effect of pulling pedestrians towards their goal position.
The term $\Vec{\mathbf{F}}_{veh}$ is used to shape the effect of the vehicle on the pedestrian motion, whereas the term $\Vec{\mathbf{F}}_{soc}$ is the so called social force. The social force models how pedestrians interact with each other but since we are mainly concerned on AV decision-making in the presence of a single pedestrian, we will neglect this term in the further discussions within this paper.

\subsection{Deep Reinforcement Learning}
%\begin{figure}[tpb]
%\centering
%    \includegraphics[width=0.30\textwidth]{RL.png}
%    \caption{The RL framework. As the agent interacts with the environment, the RL algorithm updates the policy function based on the experience gathered in order to improve the policy and achieve better cumulative reward in the future.}  
 %   \vspace{-.4 cm}
%    \label{fig:rl}
%\end{figure}

A reinforcement learning problem is formulated as a Markov Decision Process (MDP) $(S, A, \mathcal{T}, R, \gamma)$, where $S$ is the set of possible agent and environment states,
$A$ is the set of actions, $\mathcal{T}$ is the state transition probability, and $R$ is the reward function.
Two entities interact with each other at all times: an agent and the environment. The agent selects an action $a_t$ in the set $A$, which causes a change in the state $s_t$ according to the transition probability $\mathcal{T}(s_t|s_{t-1}, a_t)$ under action $a_t$. The agent is also provided with a numerical reward $r_t$, based on the outcome of the action taken.
The goal of an RL algorithm is to learn an optimal policy $\pi^*(a|s)$, which is a mapping from the state space to the action space, that maximises the expected future total reward:
\vspace{-0.1cm}
\begin{equation}
    J(\pi) = \mathbb{E}_{\pi} [ \sum_{k=0}^\infty \gamma^k r_{t+k} ]
\end{equation}
where $\gamma$ is called the discount factor. Gamma is less than 1, so events in the distant future are weighted less than events in the immediate future.

\vspace{-0.5cm}
\subsection{Soft-Actor Critic Algorithm}
The Soft-Actor Critic algorithm (SAC) \cite{haarnoja2018soft} is an RL algorithm that combines the RL framework with the principle of maximum entropy. The policy seeks to maximise a modified version of the expected future reward which is defined as:
\begin{equation}
    \max_{\pi} J(\pi) = \sum_{t=0}\mathbb{E}_{\pi}[r(s_t, a_t) + \alpha \mathcal{H}(\pi(\cdot|s_t))]
\end{equation}
$J(\pi)$ maximises both the expected cumulative reward and an entropy term $\mathcal{H}(\pi(\cdot|s_t))$, to encourage exploration at the time of training and improve training speed. The parameter $\alpha$ is known as the temperature and it affects the weight of the entropy term. 
Precisely, SAC aims to learn three functions: the policy network with parameter $\theta$, $\pi_{\theta}$, a soft Q-value function parametised by $w$, $Q_w$, and a soft state value function parametrised by $\psi$, $V_\psi$.
The experience gathered by the agent is stored in a replay buffer and, similar to DQN and DDPG, the Q network and the value network are trained using supervised learning with the data contained in a replay buffer. The targets for the network update are defined as:
\begin{equation}
\hat Q(s_t, a_t) = r(s_t, a_t) + \gamma \mathbb{E}_{s_{t+1} \sim \rho_{pi}(s)}\left[V_{\psi}(s_{t+1}) \right]
\end{equation}
\vspace{-0.6cm}
\begin{equation}
    \hat V(s_t) = \mathbb{E}_{a_t \sim \pi_\theta} \left[Q_w(s_t, a_t) - \alpha \log \pi_\theta(a_t|s_t) \right]
\end{equation}
The policy is parametrised as stochastic neural network $a_t = f_{\theta}(\epsilon_t, s_t)$, where $\epsilon_t$ is an input noise vector, sampled from a Gaussian distribution. Then objective function for policy optimization can be rewritten as:
\begin{equation}
    J_\pi(\theta) = \mathbb{E}_{s_t, \epsilon_t}\left[
    \alpha \log (\pi_\theta(f_{\theta}(\epsilon_t, s_t)|s_t) - Q_\theta(s_t, f_{\theta}(\epsilon_t, s_t))\right]
\end{equation}

\vspace{-0.5cm}
\subsection{Proximal Policy Optimisation Algorithm}

Proximal Policy Optimisation (PPO) \cite{schulman2017proximal} is a model-free Deep RL algorithm designed for continuous actions spaces. In order to improve training stability, PPO imposes a constraint on the size of the policy update at each iteration, which results in smoother policies that are appealing when considering our problem from an ergonomics perspective.

The objective function measures the total advantage over the state visitation distribution and actions. In a standard off-policy algorithm it can be expressed as:
$$J(\theta) = \mathbb{E}_{a \sim \beta} [\frac{\pi_{\theta}(a|s)}{\beta(a|s)} A_{\theta_{old}}(s, a) ] $$
where $\beta(a|s)$ is the sampling distribution.

Since PPO uses the old policy $\theta_{old}$ to generate data and we update the parameters $\theta$, the objective function becomes:
\begin{equation}
\label{eq:TRPO}
    J(\theta) = \mathbb{E}_{a \sim \pi_{\theta_{old}}} [\frac{\pi_{\theta}(a|s)}{\pi_{\theta_{old}}(a|s)} A_{\theta_{old}}(s, a) ] 
\end{equation}

PPO updates the policy parameters $\theta$ so as to maximise a slightly modified version of equation \ref{eq:TRPO}, which takes into account the constraint on the policy update, by the addition of a clipping parameter $\epsilon$.
Let $r(\theta) = \pi_{\theta}(a|s)/ \pi_{\theta_{old}}(a|s)$, the modified objective function for PPO is:
\begin{multline}
\label{eq:PPO}
    J(\theta) = \mathbb{E}_{a \sim \pi_{\theta_{old}}} [\text{min}(r(\theta) A_{\theta_{old}}(s, a), \\
    \text{clip}(r(\theta), 1 - \epsilon, 1+\epsilon) A_{\theta_{old}}(s, a)) ] 
\end{multline}
The objective of this modified objective function is to discourage policy updates that would cause big policy parameters variation even though they would lead to greater rewards.

\vspace{-0.3cm}

\subsection{Social Value Orientation}
%\begin{figure}[htpb]
%\centering
%    \includegraphics[width=0.30\textwidth]{svo.png}
%    \caption{Social Value Orientation ring. The SVO value $\varphi$ affects the behaviour of the ego-vehicle.}
%    \label{fig:svo-ring}
%    \vspace{-.6 cm}
%\end{figure}
In Social Psychology, Social Value Orientation (SVO) \cite{mcclintock1989social, van1997development} is a value that describes how much a person values other people's welfare in relation to their own. 
Each individual can be modelled as an agent that selects actions so as to maximise their own utility function. We can model each individuals social preferences by expressing their own utility function as a combination of two terms, the ego agent's selfish utility $U_{self}$ and a term that captures agents' utility $U_{other}$:

\begin{equation} \label{eq:utility}
    U_{total} = \cos(\varphi)U_{self}+ \sin(\varphi)U_{other}
\end{equation}
where $\varphi$ is the SVO value. It is an angle, whose value affects the weights of the two utility terms, and therefore the balance between selfish and altruistic rewards. 
%As highlighted in Fig. \ref{fig:svo-ring}, 
We can characterise the personality of each individual with the SVO value. For example, an SVO value of 90° corresponds to fully altruistic behaviour, whereas an SVO value of 0° corresponds to an individualistic agent. 
In our work, we focused on SVO values between 0° and 90°, as we want the AV to exhibit pro-social behaviour and yield to the pedestrian if necessary to avoid dangerous situations. 

SVO has been previously used to design controllers in a game-theoretic setting \cite{schwarting2019social}, but this demands long complex computations to solve for a Nash equilibrium points. 
In our work, we try to mitigate the computational cost of the optimisation problem by using SVO in the RL framework, thereby moving the computational cost from execution time to training time, in a learning-based fashion. 
We integrate the SVO concept directly in the MDP model formulation by constructing a reward function that is composed of two terms, one that models the AV's own objective $U_{self}$ and one that models the pedestrian's objective $U_{other}$.
%As model-free RL algorithms are computationally less complex and do not require an accurate representation of the environment to be effective, we choose the SAC algorithm, which has proven to be very effective for autonomous car traffic navigation \cite{chen2019model}. The SAC has also the advantage of using a continuous action space, which is more suitable for our problem. 
%To generate realistic and interactive experience at training and test time, we use social forces to model the pedestrian's behaviour. We adopted a model similar to the one used in \cite{yang2021sub}, where we considered a single vehicle and neglected the interaction with other pedestrians. 
%We include an awareness probability for the pedestrian to improve robustness to collisions and avoid overfitting of the pedestrian behaviour. This way, the trained policy will be able to deal with dangerous situations where the pedestrian starts crossing without seeing the vehicle. 
%In the next section, we introduce the state and action spaces of the MDP and in section XXX we describe the social reward function design with SVO.

\section{Situational Aware Pedestrian Model}
We modify the traditional social force based model by mixing it with a gap-acceptance model to simulate pedestrian crossing behaviours. In this way, we are able to obtain realistic trajectories due to the social force component while still maintaining the advantages of gap-acceptance models, i.e. the accurate description of crossing initiation. 

\vspace{-0.3cm}
\subsection{Pedestrian's Motivation}
We model the pedestrian situational awareness as a number that represents the pedestrian's willingness to cross the road, which we term \textit{motivation}. Inspired by the work of \cite{markkula2018models}, we model the pedestrian's motivation as a discrete time variable that quantifies the pedestrian's crossing willingness. The motivation takes into account environmental factors such as the AV's forward velocity $v_{v}$, the distance between the pedestrian and the vehicle $D_{pv}$, the lane width and the vehicle's acceleration perceived by the pedestrian $a$.
% The motivation should capture the decision-making process of pedestrians, in particular we model it as a dynamic variable whose value changes based on the human's understanding of the environment.

The motivation at any point in time $M(t)$ is a real value in the interval $\left[0, 1\right]$, with $1$ indicating that the pedestrian wants to cross the road and $0$ the opposite. In order to model the fact that the decision-making process is made over time, we apply a first order filter and update the motivation according to the following equation:
\begin{equation}
    M(t+1) = \alpha M(t) + (1 - \alpha) \hat{M}(t)
\end{equation}
where $\hat{M}(t)$ is an innovation term that is computed according to the vehicle's position and actions and $M(t)$ is the motivation at the previous timestep.

The innovation term is computed as a logistic function:
\begin{equation}
\label{eq:innovation}
    \hat{M}(t) = \frac{1}{1 + e^{-\left(\bm{\psi}^T\mathbf{f} - \beta \right)}}
\end{equation}
where $\mathbf{f}$ is a vector of features, $\bm{\psi}$ is a vector of weights, and $\beta$ is a parameter. The vector of features combines the advantage time and the acceleration of the vehicle perceived by the pedestrian:
\begin{equation}
    \mathbf{f} = \left[t_{adv}, a \right]^T
\end{equation}

In particular, We define the advantage time $t_{adv}$ as the difference between the time to collision and the time that the pedestrian needs to cross the road, considering their reaction time:
\begin{equation}
    t_{adv} = \frac{D_{pv}}{v_{v}} - \frac{kL}{v_d} - t_{r}
\end{equation}
where $L$ is the lane width, $k$ is a coefficient which is equal to $1.0$ if the pedestrian initiates crossing on the same side as the vehicle's lane or $2.0$ otherwise, indicating that the pedestrian has to travel only half the road width or the total road width. $t_r$ is an additional time factor that takes into account pedestrian reaction time. The terms $v_v$ and $v_d$ represent the vehicle's speed and the pedestrian desired walking speed.  

\vspace{-0.3cm}
\subsection{Navigational Force}
The navigational force is a proportional controller that drives the pedestrian towards their goal, weighted by the current pedestrian motivation:
\begin{equation}
    \Vec{\mathbf{F}}_{nav}(t) = M(t) \cdot k_d \left(\Vec{\mathbf{v}}(t) - \Vec{\mathbf{v}}_d(t) \right)
\end{equation}

The desired velocity $\Vec{\mathbf{v}}_d(t)$ points at each timestep in the direction of the goal $\Vec{\bm{g}}$ and has a magnitude equal to the pedestrian's preferred walking speed $v_d$:
\begin{equation}
    \Vec{\mathbf{v}}_d(t) = v_d \frac{\left(\Vec{\bm{g}} - \Vec{\bm{p}} \right)}{\sqrt{\Vert \Vec{\bm{g}} - \Vec{\bm{p}} \Vert^2 + \sigma^2}}
\end{equation}
where $\Vec{\bm{p}}$ is the pedestrian's current position, and $\sigma$ is a regularisation factor to avoid the problem of division by zero.

\vspace{-0.3cm}
\subsection{Vehicle Interaction}
We modelled the vehicle influence on the pedestrian as a superposition of three different force fields. The first term affects the pedestrian trajectory so that they avoid collisions with the vehicle, the second term encourages walking around the vehicle when it has very low speed, and the last term pushes the pedestrian away from the front area of the vehicle if it is approaching with high speed. Since a pedestrian will avoid walking in the area in front of a vehicle approaching at high speed and will not initiate walking around it unless the vehicle speed is sufficiently low, we take this into account by introducing a velocity coefficient that blends the second and third term according to the vehicle's speed.
In particular, we define the overall force field as:
\begin{equation}
    \Vec{\mathbf{F}}_{veh} = \Vec{\mathbf{F}}_{shape} + k(v) \Vec{\mathbf{F}}_{flow} + (1 - k(v)) \Vec{\mathbf{F}}_{speed}
\end{equation}
where
\begin{equation}
    k(v) = \frac{1}{1 + k_v v^2}
\end{equation}
The parameter $k(v)$ is used to obtain a linear combination of the fields $\Vec{\mathbf{F}}_{flow}$ and $\Vec{\mathbf{F}}_{speed}$, so that at lower velocities the former prevails, whereas at higher speeds the latter prevails. 
Let $\Vec{\mathbf{p}} = \left[x, y\right]^T$ be the coordinates of a pedestrian in the vehicle local frame.

\begin{figure}[pb]
\centering
\begin{subfigure}[t]{0.20\textwidth}
    \includegraphics[width=\linewidth]{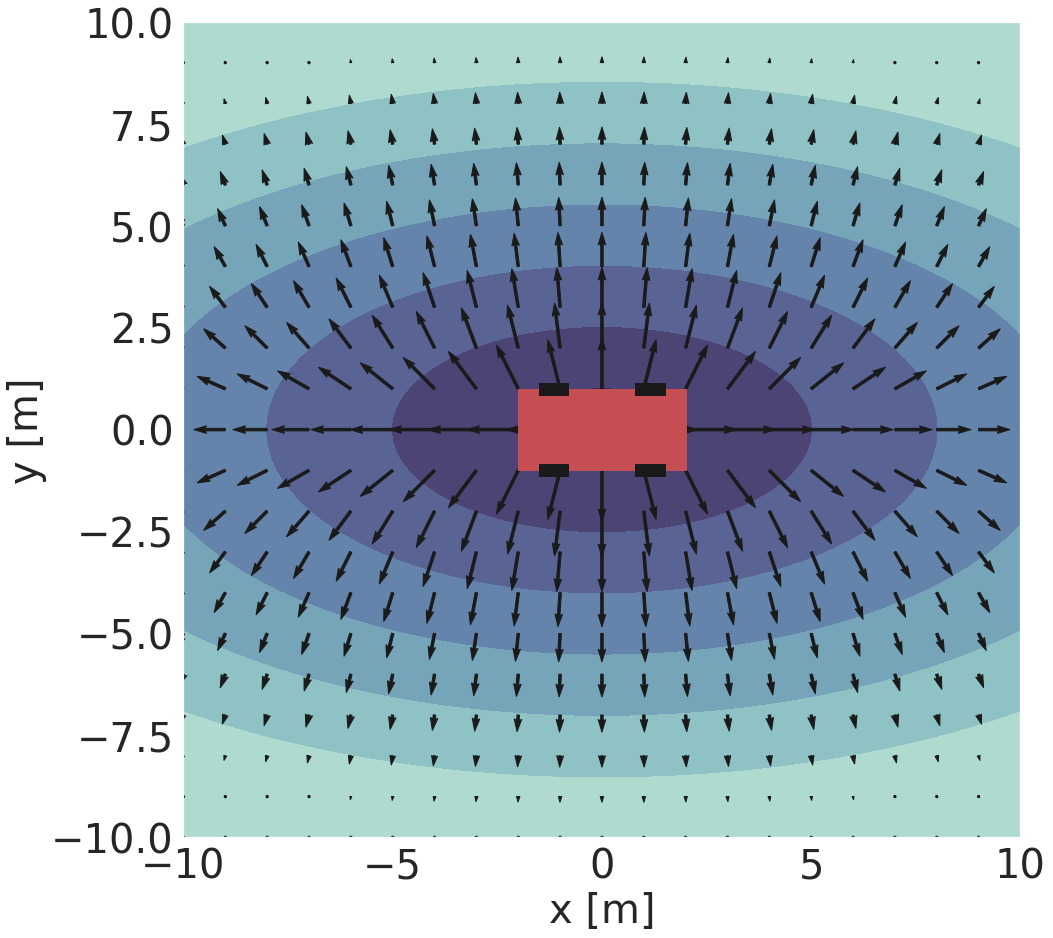}
    \end{subfigure}
\begin{subfigure}[t]{0.20\textwidth}
    \includegraphics[width=\linewidth]{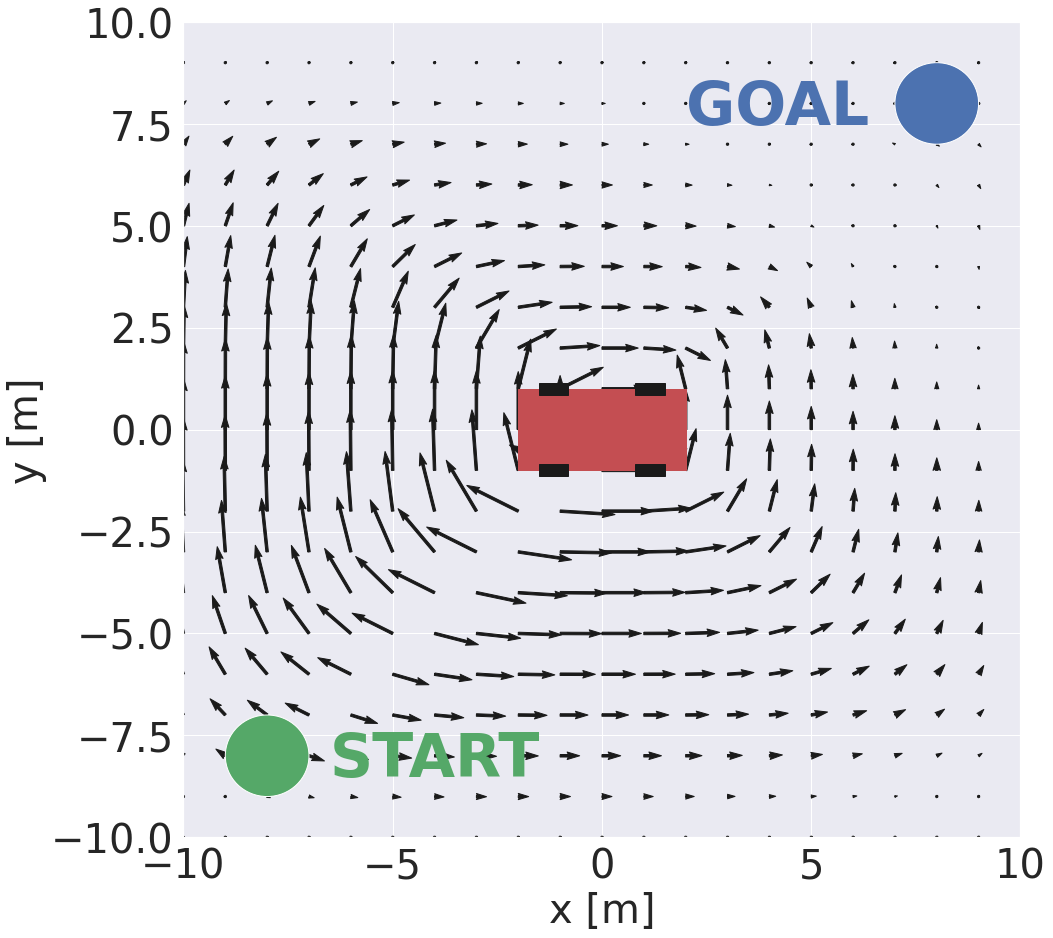}
    \end{subfigure}
\caption{Shape field (left) and force field (right) representation. The flow field is shown with two randomly chosen start and goal positions.}
\label{fig:flows}
\end{figure}

The shape of the fields $\Vec{\mathbf{F}}_{shape}$ and $\Vec{\mathbf{F}}_{flow}$ is shown in Fig. \ref{fig:flows}.
We approximate the AV shape as an ellipsis for the sake of the repulsive force modelling, with semi-axes $a$ and $b$, equal to half the vehicle length and width respectively. 

\begin{figure}[tb]
\centering
    \includegraphics[width=0.48\textwidth]{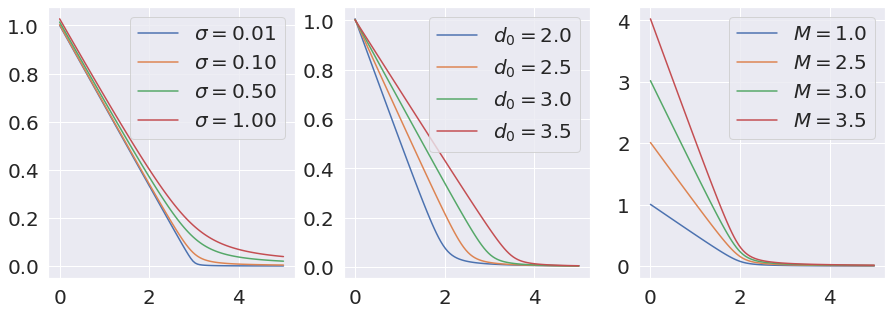}
    \caption{Linear decay with smoothing. Values are unitless.}
    \label{fig:linear_decay}
    \vspace{-0.6 cm}
\end{figure}

We use a linear decay function with smoothing to model the influence of the vehicle shape on the pedestrian based on the distance $d$ between the vehicle and the pedestrian, which is defined as:
\vspace{-0.1cm}
\begin{equation}
    h(d; A, d_0, \sigma) = \frac{A}{2d_0}\left(d_0 - d + \sqrt{\left(d_0 - d\right)^2 + \sigma} \right)
\end{equation}
\vspace{-0.1cm}
where $A$, $d_0$, and $\sigma$ are parameters that determine the shape of the linear decay function 
and whose effects are shown in Fig. \ref{fig:linear_decay}. 
We use an elliptical distance in accordance to the AV's shape approximation:
\vspace{-0.1cm}
\begin{equation}
    d = \sqrt{\left(\frac{x}{a} \right)^2 + \left(\frac{y}{b} \right)^2}
\end{equation}
\vspace{-0.1cm}
where $x$ and $y$ are the pedestrian's coordinates relative to the vehicle.

The repulsive force direction is orthogonal to the vehicle's shape approximation and its magnitude depends on a linear decay with smoothing function of the elliptical distance. The denominator normalises the equation to get a unit vector in the desired direction:
\begin{equation}
    \Vec{\mathbf{F}}_{shape} = \frac{h(d; A_{s}, d_{0s}, \sigma_s)}{\sqrt{\left(\frac{2x}{a^2}\right)^2 + \left(\frac{2y}{b^2}\right)^2}}
    \left(\frac{2x}{a^2}\hat{\boldsymbol{\imath}} + \frac{2y}{b^2} \hat{\boldsymbol{\jmath}}\right)
\end{equation}

The flow field encourages the pedestrian to walk around the vehicle.
We introduce a coefficient $k_{f}(\Vec{\bm{p}})$ which has two purposes. Its sign determines if the pedestrian walks around the vehicle clockwise or counterclockwise and it is decided by estimating the shortest path between the pedestrian current position and the goal position.
The magnitude of the coefficient $k_{f}(\Vec{\bm{p}})$ is one at the beginning of the trajectory and decreases to zero as the pedestrian is closer to the goal. This choice was made because the vehicle should not influence the pedestrian motion once the pedestrian has passed the vehicle and is moving further away from it. In symbols:
\begin{equation}
    \left|k_{f}(\Vec{\bm{p}})\right| = 
\begin{cases}
    1.0& \text{if } P < 0\\
    0& \text{if } P > \Vert \Vec{\bm{g}} - \Vec{\bm{p}}_0 \Vert\\
    \frac{\Vert \Vec{\bm{g}} - \Vec{\bm{p}_0}\Vert - P}{\Vert \Vec{\bm{g}} - \Vec{\bm{p}_0}\Vert}& otherwise
\end{cases}
\end{equation}
where $P$ is the pedestrian progress towards their goal:
\begin{equation}
    P = \frac{\left(\Vec{\bm{p}} - \Vec{\bm{p}_0}\right)\cdot 
    \left(\Vec{\bm{g}} - \Vec{\bm{p}_0}\right)}{\Vert \Vec{\bm{g}} - \Vec{\bm{p}_0}\Vert}
\end{equation}

We allow the flow term to have its own linear decay with smoothing parameters.
Compared to the shape field, the flow field has a different power for the $x$ and $y$ terms in eq. \ref{eq:flow_field} to make the pedestrian trajectory follow more realistic paths around the vehicle. The negative sign in the last term of equation \ref{eq:flow_field} makes the field rotate around the vehicle.
The pedestrian flow term can then be defined as:
\begin{equation}
\label{eq:flow_field}
    \Vec{\mathbf{F}}_{flow} = k_{f}(\Vec{\bm{p}})\frac{h(d; A_{f}, d_{0f}, \sigma_f)}{\sqrt{\left(\frac{-2y^3}{b}\right)^2 + \left(\frac{2x^3}{a}\right)^2}}
    \left(\frac{-2y^3}{b}\hat{\boldsymbol{\imath}} + \frac{2x^3}{a} \hat{\boldsymbol{\jmath}}\right)
\end{equation}

The influence of the vehicle speed on the pedestrian motion is modelled with the force field $\Vec{\mathbf{F}}_{speed}$, which follows an exponential decay:
\begin{equation}
    \Vec{\mathbf{F}}_{speed} = A \cdot \sign{(y)} \exp{\left(-\frac{x - a}{v\Delta T}\right)}
    \exp{\left(-\frac{y^2}{2\sigma_y^2}\right)} \hat{\boldsymbol{\jmath}}
\vspace{-.1cm}
\end{equation}

where $A$ is a scaling coefficient, $v_v$ represents the vehicle speed, $\Delta T$ is a time factor, and $\sigma_y$ is a constant proportional to the lane width. The exponential decay is influenced by the vehicle speed, varying the length of the area that is influenced in front of the AV.

{\tiny
\begin{algorithm}[ptb]
\caption{Pedestrian model.}\label{alg:alg1}
\begin{algorithmic}
\STATE 
\STATE {\textbf{Input:}} Pedestrian's position $\Vec{\bm{p}} = \left[x, y\right]^T$, pedestrian's goal position $\Vec{\bm{g}}$, vehicle speed $v$ and acceleration $a$.
\STATE {\textbf{Output:}} Pedestrian acceleration $\Vec{\bm{a}}$ 
\STATE {\textbf{Parameters:}} see Table \ref{tab:parameters}
%\STATE
\STATE {\textbf{Compute Pedestrian Speed:}}
\STATE \hspace{0.5cm} \textit{Update Motivation:}
\STATE \hspace{1.0cm} $\hat{M}(t) \gets \frac{1}{1 + e^{-\bm{\psi}^T\mathbf{f}}}$
\STATE \hspace{1.0cm} $M(t) \gets \alpha M(t-1) + (1-\alpha) \hat{M}(t)$

\STATE \hspace{0.5cm} \textit{Update Forces:}
\STATE \hspace{1.0cm} \textbf{if } $M(t) > \theta_f$:
\STATE \hspace{1.5cm} $
    \Vec{\mathbf{F}}_{nav}(t) \gets M(t) \cdot k_d \left(\Vec{\mathbf{v}}(t) - \Vec{\mathbf{v}}_d(t) \right)$
\STATE \hspace{1.0cm} \textbf{else: }
\STATE \hspace{1.5cm} $\Vec{\mathbf{F}}_{nav}(t) \gets \bm{0}$
\STATE \hspace{1.0cm} $\Vec{\mathbf{F}}_{sh}(t), \Vec{\mathbf{F}}_{f}(t), \Vec{\mathbf{F}}_{sp}(t) \gets$ \textit{Update Forces}
\STATE \hspace{1.0 cm} $
    k(v) \gets \frac{1}{1 + k_v v^2}$
\STATE \hspace{1.0cm} $\Vec{\mathbf{F}}_{veh}(t) \gets \Vec{\mathbf{F}}_{sh}(t) + k(v)\Vec{\mathbf{F}}_{f}(t) + (1-k(v) \Vec{\mathbf{F}}_{sp}(t)$
\STATE \hspace{1.0cm} $\Vec{\mathbf{F}}_{tot}(t) \gets \Vec{\mathbf{F}}_{nav}(t) + \Vec{\mathbf{F}}_{veh}(t)$
\STATE \hspace{0.5cm} \textit{Determine acceleration:}
\STATE \hspace{1.0cm} $\Vec{\bm{a}}(t) \gets \Vec{\mathbf{F}}_{tot}(t) / m$
\STATE \hspace{1.0cm} \textbf{if} $\Vert \Vec{\bm{a}}(t) \Vert > a_{max}$:
\STATE \hspace{1.5cm} $\Vec{\bm{a}}(t) \gets a_{max} \cdot \Vec{\bm{a}}(t)/\Vert \Vec{\bm{a}}(t) \Vert$
\STATE \hspace{1.0cm} $\Vec{\bm{v}}(t) \gets \Vec{\bm{v}}(t-1) + \Vec{\bm{a}}(t) \cdot T_s$
\STATE \hspace{1.5cm} $\Vec{\bm{v}}(t) \gets v_{max} \cdot \Vec{\bm{v}}(t)/\Vert \Vec{\bm{v}}(t) \Vert$
\end{algorithmic}
\label{alg1}
\end{algorithm}
\vspace{-.4cm}
}
\begin{table}
\begin{center}
\caption{Parameter Set.}
\label{tab:parameters}
\begin{tabular}{| c | c | c |}
\hline
Type & Parameter name & Values\\
\hline
Motivation& $\alpha, v_d, t_r, \bm{\psi}, \theta_f, \beta$ & $(0.8, 2.0, 0.05, [3.0, -0.3],$ \\
& & $0.3, 2.2)$\\
\hline
Navigation& $k_d, \sigma_d$ & $\left(200, 0.09 \right)$\\
\hline
Shape& $M_s, d_{0s}, \sigma_s$ & $\left(800, 4.0, 0.1 \right)$\\ 
\hline
Flow& $M_s, d_{0f}, \sigma_f$ & 
$\left(600, 6.0, 0.1 \right)$\\
\hline 
Speed& $A, \Delta T, \sigma_y$ & 
$\left(400, 1.0, 0.2L \right)$\\
\hline 
Constraints& $a_{max}, v_{max}, m, k_v$ & $\left(3.0, 4.0, 75, 0.1 \right)$\\
\hline 
\end{tabular}
\end{center}
\end{table}

\section{SVO-informed Vehicle Controller}
We design a Deep Reinforcement Learning environment in which the Autonomous Vehicle (the RL agent) interacts with the pedestrian. We let the AV learn its behavioural policy by experiencing interactions with the pedestrian model we developed. We use two different DRL algorithms, SAC and PPO, to train two sets of  policies and compare their performances.  
We will now describe how we model the pedestrian collision-avoidance problem as a Markov Decision Process (MDP) that can be used to train DRL AV agents.

\vspace{-0.3cm}
\subsection{MDP Formulation}
\label{sec:mdp}
\subsubsection{State Space}
\hfill\\
In our model, we focused on a scenario consisting of a straight lane and a single pedestrian (see Fig. \ref{fig:scene}). 
We assume the ego-vehicle is able to locate itself with respect to a reference path generated by a global routing module. The purpose of the policy network output is to control the vehicle along the planned trajectory taking the pedestrian's behaviour into account. We also assume that a perception module is available to locate the pedestrian's pedestrian position relative to the vehicle.
Therefore, the observation space $s_t$ available to the ego-vehicle consists of (1) the vehicle longitudinal velocity $v^{ego}_t$, (2) the pedestrian relative position $\mathbf{p}_t$ and velocity $\mathbf{v}^{ped}_t$:
\begin{equation}
    s_t = [v^{ego}_t, \mathbf{p}_t,  \mathbf{v}^{ped}_t]^T  \in \mathbb{R}^5
\end{equation}

\subsubsection{Action Space}
The action space used consists of the longitudinal acceleration of the vehicle $a_t$. The normalised output of the policy network is then scaled into the interval $\left[-0.3\textit{g}, 0.3\textit{g}\right]$.

\vspace{-0.3cm}
\subsection{Social Reward Function}
\label{sec:reward}
The designed reward function consists of two terms:
\begin{equation}
\label{eq:reward}
r(s_t,a_t) = \cos{\varphi}\cdot r_{car}(s_t, a_t) + \sin{\varphi} \cdot r_p(s_t, a_t)
\end{equation}
where $r_{car}$ indicates a reward function that takes the vehicle own performance parameters into account and the $r_p$ term is a term that captures the pedestrian's intentions and comfort. $\varphi$ is the ego-vehicles SVO value, which is used to shape the car's altruistic or egoistic behaviour.

The first term $r_{car}$ in the reward function is also a combination of multiple terms:
\begin{equation}
    r_{car}(s_t, a_t) = r_c + r_g + r_v
\end{equation}
where $r_c$ is a penalty in case of collision, $r_g$ is the reward for reaching the goal, and $r_v$ is a speed reward that encourages the AV to complete the task as quickly as possible. Empirically, we set $r_c = -100$, $r_g = 40$, and $r_v = -4$.

The second term of equation \ref{eq:reward} is used to capture the pedestrian's intentions and comfort in the AV's decision-making process. We assume a pedestrian crossing the road is attempting to reach their goal in the least amount of time possible, so we give a positive reward proportional to the pedestrian crossing speed to the AV when the pedestrian is crossing. Also, since we want our RL agent to behave pro-socially, i.e. yielding to the pedestrian if necessary, we give a positive reward only if the pedestrian is crossing in front of the vehicle. Since an AV stopping in close proximity of a pedestrian to let them cross could be potentially dangerous or could make the pedestrian feel unsafe, we weight the reward by a factor $\sigma \left(D_{pv} \right)$, where $D_{pv}$ is the distance between the vehicle and the pedestrian. $\sigma \left(D_{pv} \right)$ is a sigmoid function that tends to 0 when $D_{pv}$ tends to 0.  If $\mathbf{v}_p$ is the pedestrian velocity, we can then express the pedestrian reward function as:
\begin{equation}
    r_p= 
\begin{cases}
    k_p \sigma(D_{pv}) \Vec{\bm{v}_p} \cdot \hat{\rho},& \text{if } \textit{wants to cross} \quad \textbf{and} \quad x_p > x_v\\
    0,& \text{otherwise}
\end{cases}
\end{equation}
where $k_p$ is a scaling coefficient for the pedestrian reward, $\Vec{\bm{v}_p}$ is the pedestrian velocity, and $\hat{\rho}$ is a unit vector pointing from the pedestrian to their goal position. 

\subsection{Neural Network Implementation}
The neural network architecture is the same for both PPO and SAC and consists of two fully connected layers with 256 hidden neurons each, shared by both the actor and critic networks. A simple fully-connected multi-layer perceptron network was used, as the input space is simple enough to allow us to use a simpler neural network rather than a Convolutional Neural Network which would be harder to train.

We observed that by directly training the RL agent with the new pedestrian model resulted in only aggressive policies for the RL agent, even for  SVO values close to 90°. The DRL algorithm used to get stuck in a local minimum, which caused limited exploration: since the newly introduced pedestrian model has a more cautious behaviour compared to our previous work \cite{luca21humancentric}, the AV agent optimised only the first term of eq. \ref{eq:reward}, neglecting the pedestrian's reward. We solved this issue by splitting the training in two parts. For the half of the training, the model is trained with a reckless pedestrian model that always crosses the road. For the second half of the training, we switch the pedestrian model to the more complex one. The idea behind this is that the RL model is more cautious when the conservative pedestrian is introduced, which allows to explore braking actions without falling into a local maximum for the reward. In this way, agents with higher SVO values learn that breaking yields to higher altruistic rewards.

\section{Experimental Results} 
\label{sec:experiments}
We developed a 2-D driving simulator that was used to train and subsequently test our RL agent. The simulator was developed with Python and was wrapped inside an OpenAI Gym Environment \cite{brockman2016openai}, which is a widely used interface for RL environments. We used the DRL-library Stable-Baselines3 \cite{haarnoja2018soft} for the RL algorithms, which offers implementation of many widely used RL algorithms. 
% The driving simulator implements a bicycle model for the vehicle and we model the pedestrian behaviour with algorithm \ref{alg:alg1}. 
We used a machine with one NVidia GeForce GTX 1080 Ti and a Intel(R) Core(TM) i5-6400 CPU @ 2.70GHz processor to perform the Neural Network Training. 

%\color{blue}
We divide the experimental results section as follows:
sections \ref{ssec: gap-acceptance} and \ref{ssec:qualitative-ped} present qualitative and quantitative evaluations of our pedestrian model. Section \ref{ssec:rl-scenario} introduces the reinforcement learning scenarios, section \ref{ssec:network-training} gives details about the DRL training, and section \ref{ssec:mutual-interaction} presents the evaluation of the trained agent in the interactive environment.

\subsection{Gap-Acceptance Validation}
\label{ssec: gap-acceptance}

\begin{figure}[tb]
\centering
    \includegraphics[width=0.8\linewidth]{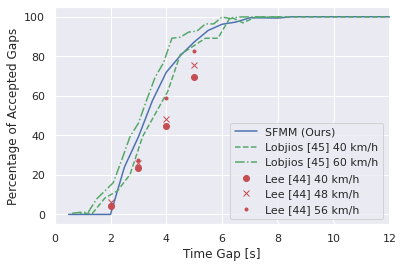}
    \caption{\color{black} Road-crossing probability in Lee et al.'s data \cite{lee2022learning} (red) and Lobjois et al.'s data \cite{lobjois2007age} (green), together with our model's (blue).}
    \label{fig:gap_acceptance}
    \vspace{-.6 cm}
\end{figure}

We use two real-world pedestrian datasets \cite{lobjois2007age, lee2022learning} to evaluate our gap-acceptance model based on motivation. 
Lee et al. \cite{lee2022learning} gathered data as part of a virtual reality experiment to investigate how the combination of kinematic information from a vehicle (e.g., Speed and Deceleration), and eHMI designs, play a role in assisting the crossing decision of pedestrians. 
The authors of \cite{lobjois2007age} designed a gap acceptance task to investigate the relationship between age difference and accepted gaps. Since age differences is not in focus in this paper, we only used the data from the age group 20-30, similar to the age range of
participants in \cite{lee2022learning}. 
In Fig. \ref{fig:gap_acceptance}, we show the gap acceptance curve generated by our Social Force Motivation model (SFMM) is in line with the empirical data and is overall capable of capturing both of this datasets well.

\vspace{-0.3cm}
\subsection{Qualitative pedestrian motion analysis}
\label{ssec:qualitative-ped}
We test the pedestrian behavioural model in the following scenarios:

\begin{itemize}
    \item fixed AV position lateral interaction;
    \item fixed AV position frontal interaction;
    \item slow-speed AV (1-5 m/s) lateral interaction;
    \item medium-speed AV  (10-15 m/s), with three different acceleration values with lateral interaction;
\end{itemize}

\begin{figure}[pb]
\centering
%\begin{subfigure}[t]{0.24\textwidth}
    %\includegraphics[width=\linewidth]{start_bottom.png}
    %\end{subfigure}
%\begin{subfigure}[t]{0.24\textwidth}
    %\includegraphics[width=\linewidth]{start_top.png}
    %\end{subfigure}
\begin{subfigure}[t]{0.24\textwidth}
    \includegraphics[width=\linewidth]{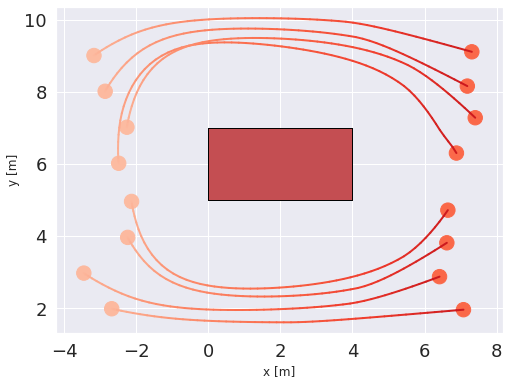}
    \end{subfigure}
\begin{subfigure}[t]{0.24\textwidth}
    \includegraphics[width=\linewidth]{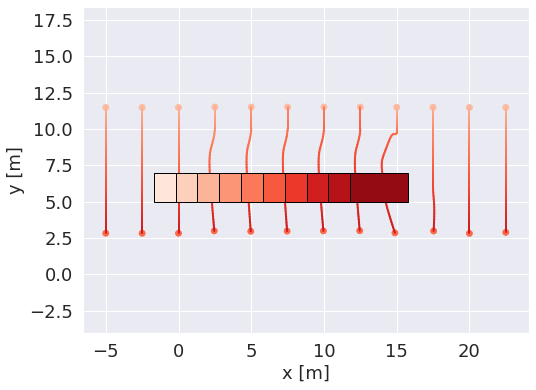}
    \end{subfigure}
\caption{Simulation trajectories. (a) Fixed AV frontal interaction crossing from bottom to top, (b) fixed AV crossing from top to bottom, (c) lateral interaction, (d) slow-moving AV. For each figure, a darker colour indicates a later simulation time. The initial position and goal positions are represented by an orange and a purple circle respectively. }
\vspace{-0.5cm}
\label{fig:ped_trajectories}
\end{figure}

We focused more on the lateral interactions between the vehicle and the pedestrian as we are mostly interested in pedestrian crossing behaviour. For each of the above scenario classes, we performed an evaluation with the pedestrian crossing from both road sides of the road.
\begin{figure}[tb]
\centering
    \includegraphics[width=0.9\linewidth]{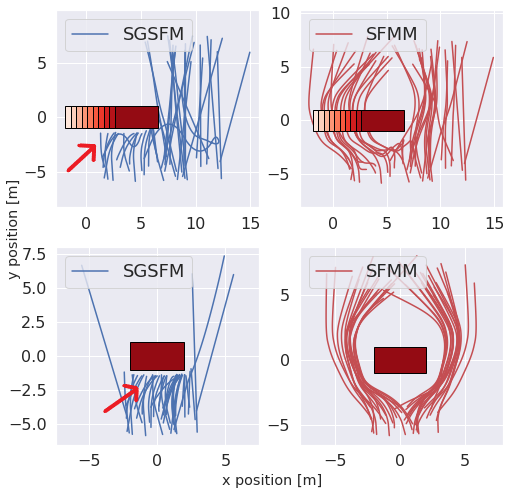}
    \caption{\color{black} Qualitative trajectory comparison between our model (red) and the model in \cite{yang2021sub} (red). We can see how our pedestrian model is capable of overcoming a static car obstacle whereas the Sub-Goal Social Force Model (SGSFM) gets stuck on opposite side of the car with respect to its goal (as indicated by the arrows). The pedestrian is trying to cross from bottom (negative y values) to the top.}
    \label{fig:traj-comparison}
    \vspace{-.6 cm}
\end{figure}
In Fig. \ref{fig:traj-comparison} we report qualitative comparative analysis of the trajectories generated by our model (red) and a state-of-the-art social force model \cite{yang2021sub} (blue).  Trajectories are obtained by changing the pedestrian spawn and goal positions, while keeping the same initial conditions for the AV. Our simulations show that the introduction of the $\Vec{\mathbf{F}}_{flow}$ term allows the pedestrian to overcome situations in which the repulsive force $\Vec{\mathbf{F}}_{shape}$ cancels out the navigational force $\Vec{\mathbf{F}}_{nav}$, allowing the pedestrian to overcome a static obstacle. This feature was not present in our previous paper \cite{luca21humancentric}. 
Fig. \ref{fig:ped_trajectories} shows additional qualitative trajectories pedestrian trajectories obtained in a frontal interaction (Fig. \ref{fig:ped_trajectories}(a)) and with car medium-speed (Fig. \ref{fig:ped_trajectories}(b)). 
Additionally, we perform a computational analysis of the pedestrian model. The results show that our model computes the pedestrian acceleration in
0.36 +- 0.3 ms against 
60 +- 2 ms for the SGSFM model \cite{yang2021sub}. Good computational performances enable faster DRL training.

\vspace{-0.1cm}
%In the trajectories generated when the pedestrian spawns close to the vehicle's front, we can see how the pedestrian tries to minimise their travel time by walking around the back of the vehicle. 
\color{black}
\vspace{-0.3cm}
\subsection{Reinforcement Learning Scenario}
\label{ssec:rl-scenario}

We trained and subsequently tested our DRL agent on a straight road scenario with a single pedestrian, (see Fig. \ref{fig:scene}).

We modelled a straight road with the ego-vehicle and a pedestrian.
We chose a road length of 60 m and width of 6 m, which is the average road width for a two lane urban road in the UK. 
The pedestrian can spawn either on the top pavement or on the bottom pavement, whereas the AV always spawns in the bottom lane. This choice does not constitute any loss of generality as we formulate the decision-making problem in the ego-vehicle reference frame.
The AV's initial speed is chosen with a uniform distribution in the interval (0 m/s, 15 m/s) as we are interested in studying a low-speed urban scenario. Selecting random values for the initial conditions favours exploration in the early stages of the training. The pedestrian's initial position along the pavement is sampled from a uniform distribution. The pedestrian's goal position is always on the opposite side of the road from their spawn point and is sampled from a normal distribution with mean value equal to the pedestrian position to ensure the distance from the crossing point is not excessive. 

\vspace{-0.3cm}
\subsection{Network Training}
\label{ssec:network-training}

\begin{figure}[t]
\centering
\begin{subfigure}[t]{0.24\textwidth}
    \includegraphics[width=\linewidth]{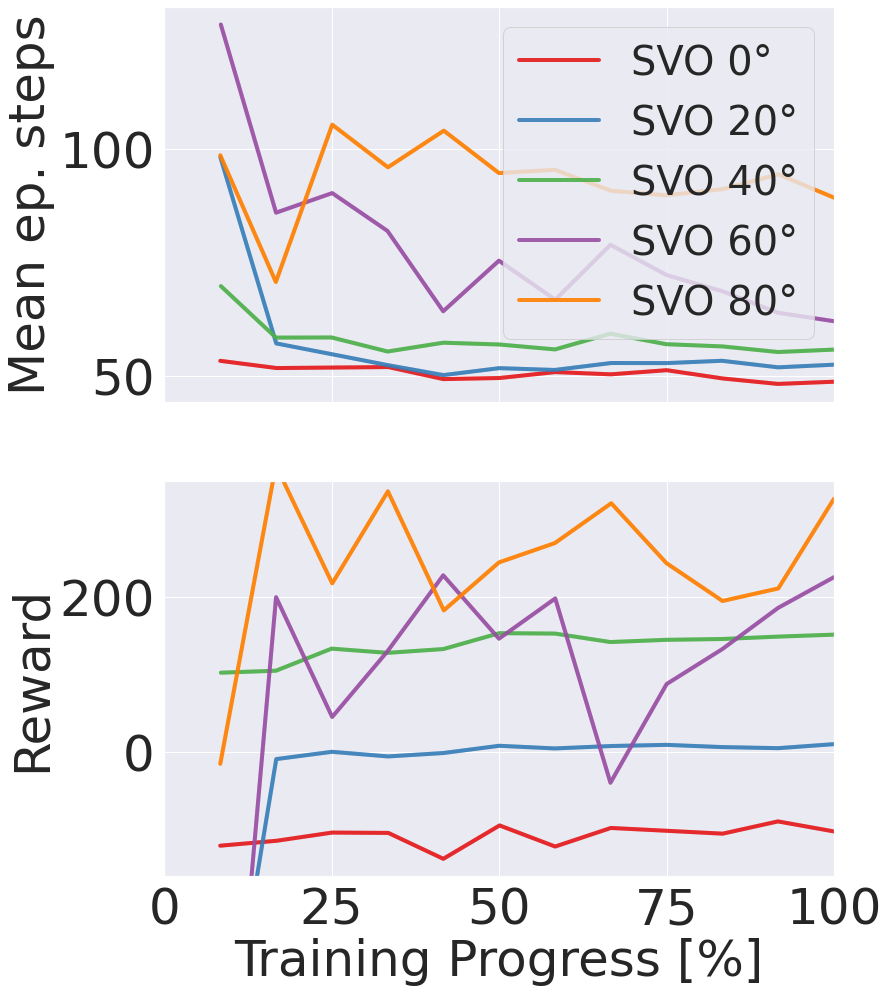}
    \end{subfigure}
\begin{subfigure}[t]{0.24\textwidth}
    \includegraphics[width=\linewidth]{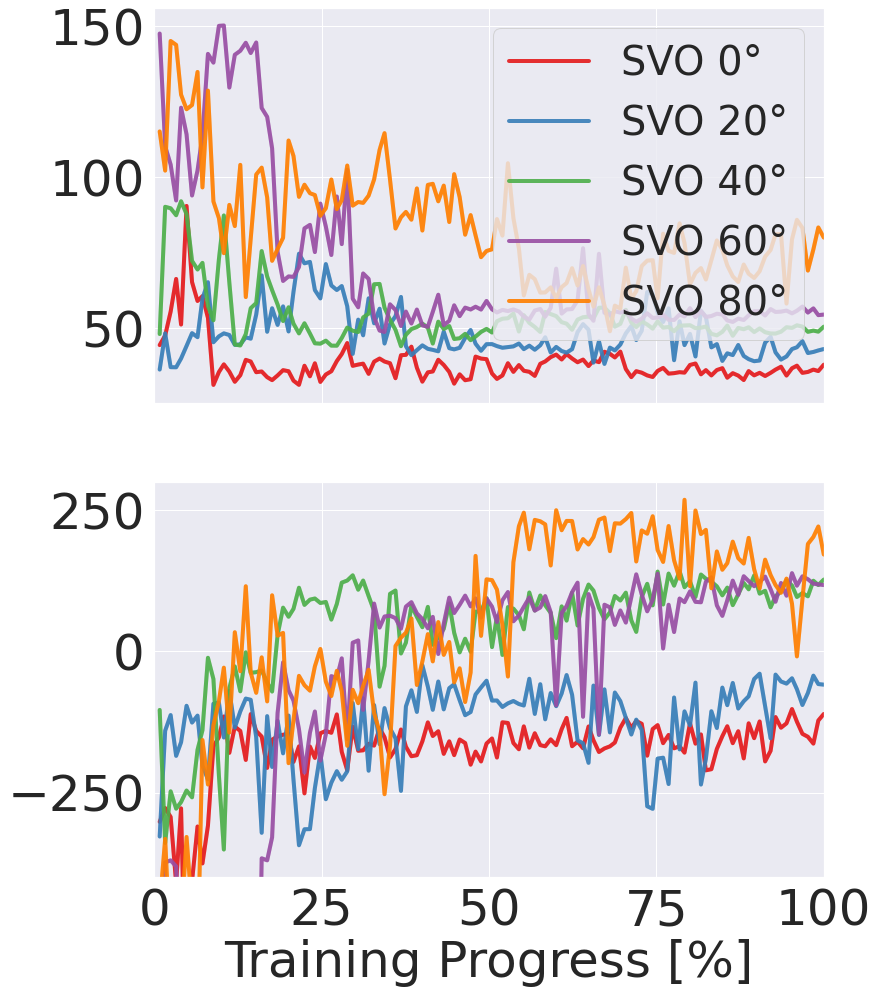}
    \end{subfigure}

\caption{Mean episode length in timesteps (top) and mean reward (bottom) with SAC algorithm (left) and PPO (right).}
\label{fig:training}
\vspace{-0.4cm}
\end{figure}

We compare the performances of two different RL algorithmsby training a total of 10 different policies: 5 for the SAC algorithm and 5 for the PPO algorithm with SVO values of 0°, 20°, 40°, 60°, and 80° respectively.
In general, the SAC algorithm requires longer than PPO to train a policy that yields the same cumulative reward, but requires fewer steps.
We compared the two algorithms by keeping the total computation time constant. We trained each policy for roughly 150 minutes, which resulted in a total of \num{2.5e06} steps for PPO and \num{2.5e05} steps for SAC. A normally distributed action noise is also added to the actions taken by the agent during training time to favour exploration. We set the replay buffer size for the SAC algorithm equal to the number of training steps so that the entire experience gathered by the agent is used during training. 
We choose a linear decay for the learning rate, initially set to \num{3e-04}. The discount factor $\gamma$ was set to $0.99$.

We show the training curves for both PPO and SAC in Fig. \ref{fig:training}. The figures show the mean episode length and the total reward gathered for different SVO values. We observe how at the end of the training the two algorithms return comparable results both in terms of mean episode length and reward gathered, which shows consistency between training instances. However, we note that for some of the SAC policies, the reward is not entirely stable at the end of the training, indicating that the SAC algorithm is much more time consuming than PPO. Nonetheless, the SAC policies yield acceptable results in terms of policy behaviour, allowing for comparison of the two algorithms.

\begin{figure}[tb]
\centering
    \includegraphics[width=1.0\linewidth]{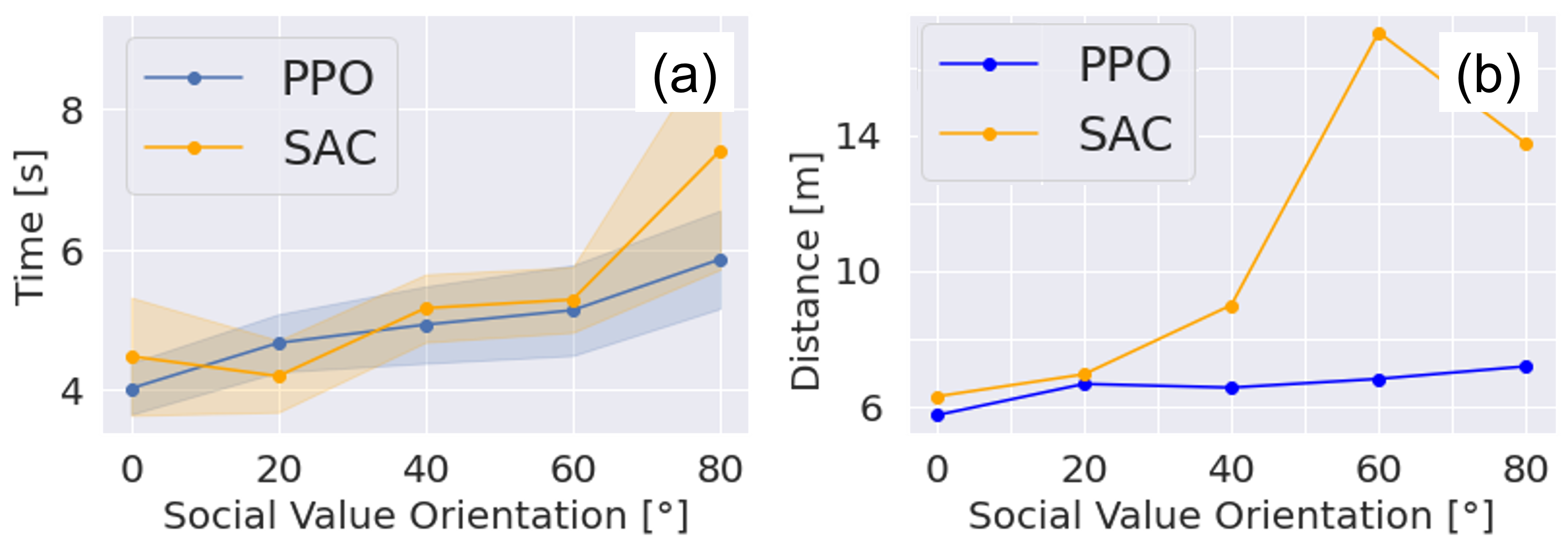}
    \caption{(a) Average time to complete task for PPO and SAC algorithms for models with different social value orientation. (b) Average minimum distance between the vehicle and the pedestrian at testing. }
    \label{fig:results}
    \vspace{-.6 cm}
\end{figure}

\vspace{-0.3cm}
%\subsection{Simulations}
\color{black}
\subsection{Mutual Interaction Evaluation}
\label{ssec:mutual-interaction}
We create two test suites of 1000 testing episodes to  evaluate the effect that our SVO reward design has on the agent behaviour. In half of the episodes, the pedestrian crosses the road from top to bottom and in the other half from bottom to top. 
The first one is used to evaluate the agent with our pedestrian model. In the second one, we increase scenario complexity by making the pedestrian unaware of the vehicle's presence, i.e. crossing regardless of the vehicle's position and speed. In this way, we were able to include hazardous and unexpected scenarios that will stress the controller robustness to the pedestrian model.
%We use the same test suite to test each RL agent under the same initial conditions. 
We analyse the smoothness of the agent trajectory, how its behaviour is affected by SVO, and the agent's robustness to the pedestrian model. 
Agents with an SVO value of 0° serve as a baseline for State of the Art DRL methods with traditional reward functions that only take the ego-vehicle's goal into account, as an agent with an SVO value of 0° is exactly equivalent to a standard DRL agent. 
\begin{figure*}[htpb]
  \centering
  \includegraphics[width=1\linewidth]{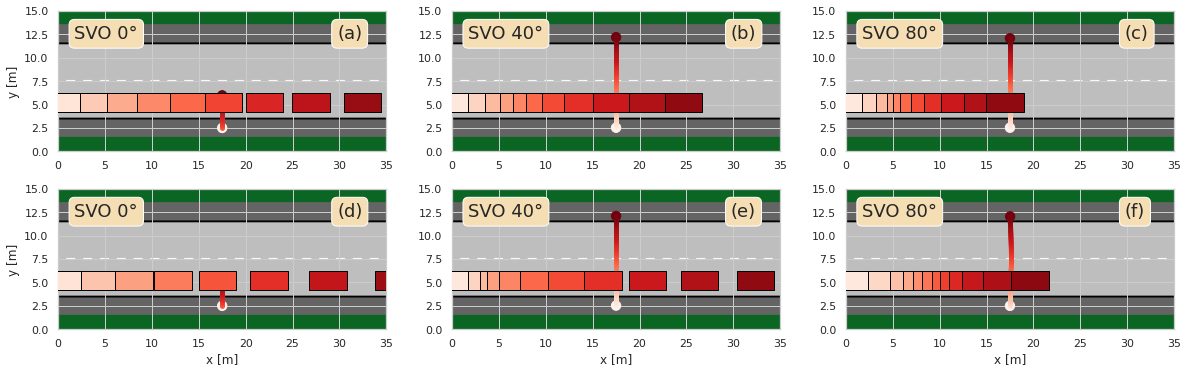} 
  \vspace{-0.5cm}
\caption{Pedestrian and vehicle agent trajectories for two episodes and three SVO values. Fig. (a)-(c) are generated with SAC and (d)-(f) with PPO. The temporal progression is indicated by coloring the car and pedestrian's trajectories from lighter to darker colors. In Fig. (b), (c), (e) and (f) the AV yields to the pedestrian, whereas in (a), (d) the pedestrian crosses after the AV has passed and has not completed crossing when the episode terminates. We can see that the 80° SVO has a less aggressive behaviour than 0° and 40°.}
\label{fig:trajectories}
\vspace{-0.5cm}
\end{figure*}

\subsubsection{Qualitative Results}
In Fig. \ref{fig:trajectories} we show pedestrians and AV trajectories with different SVO values with the same initial conditions. Agents trained with SAC (first row) and PPO (second row) display similar trajectories. In Fig. \ref{fig:trajectories}(a) and (d) the vehicle accelerates to prevent the pedestrian from crossing due to a low SVO value. Viceversa, in Fig. \ref{fig:trajectories}(c) and (f) the ego-vehicle displays a behaviour called early-stopping, in which it slows down to let the pedestrian initiate crossing. Fig. \ref{fig:trajectories}(b) and (e) have intermediate behaviour. The effects of the SVO with the overall agent behaviour are in line with our expectations, i.e. pro-social behaviour for high SVO values and egoistic behaviour with low SVO. 

\begin{figure}[pbth]
\centering
    \includegraphics[width=1.0\linewidth]{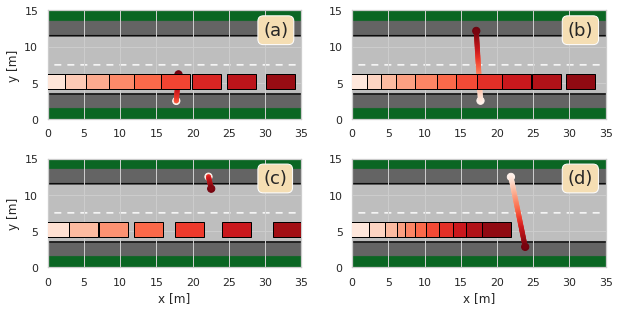}
    \caption{\color{black} Qualitative trajectories with unaware pedestrian (b), (d) and aware pedestrian (a), (c). Figures on the same row share the same initial conditions. The ego-vehicle agent is the same for all scenarios (SVO 0°) and is capable of distinguishing exploitable pedestrian behaviours from hazardous ones.}
    \label{fig:aggr_ped}
    \vspace{-.4 cm}
\end{figure}

Fig. \ref{fig:aggr_ped} shows the qualitative effect that unpredictable pedestrian behaviour has on an agent with SVO 0°. Fig. \ref{fig:aggr_ped}(a) and (c) have an aware pedestrian, Fig. \ref{fig:aggr_ped}(b) and (d) an unaware pedestrian. Despite the fact that the controller SVO is 0°, the car stops to let the pedestrian cross in order to avoid collision, thereby favouring safety over its own egoistic behaviour. 

\subsubsection{Quantitative Results}
\label{sec:quant_results}
First of all, we evaluate the agents success rate in completing its task in the first and second test suites. We consider an episode successful when the ego-vehicle reaches the end of the road whilst avoiding the pedestrian. All the agents successfully completed the task without collisions with the pedestrian in both the first and second test suite, which demonstrates the fact that our model is capable of handling the added complexity of risky scenarios.

\begin{figure}[pbth]
\centering
    \includegraphics[width=1.0\linewidth]{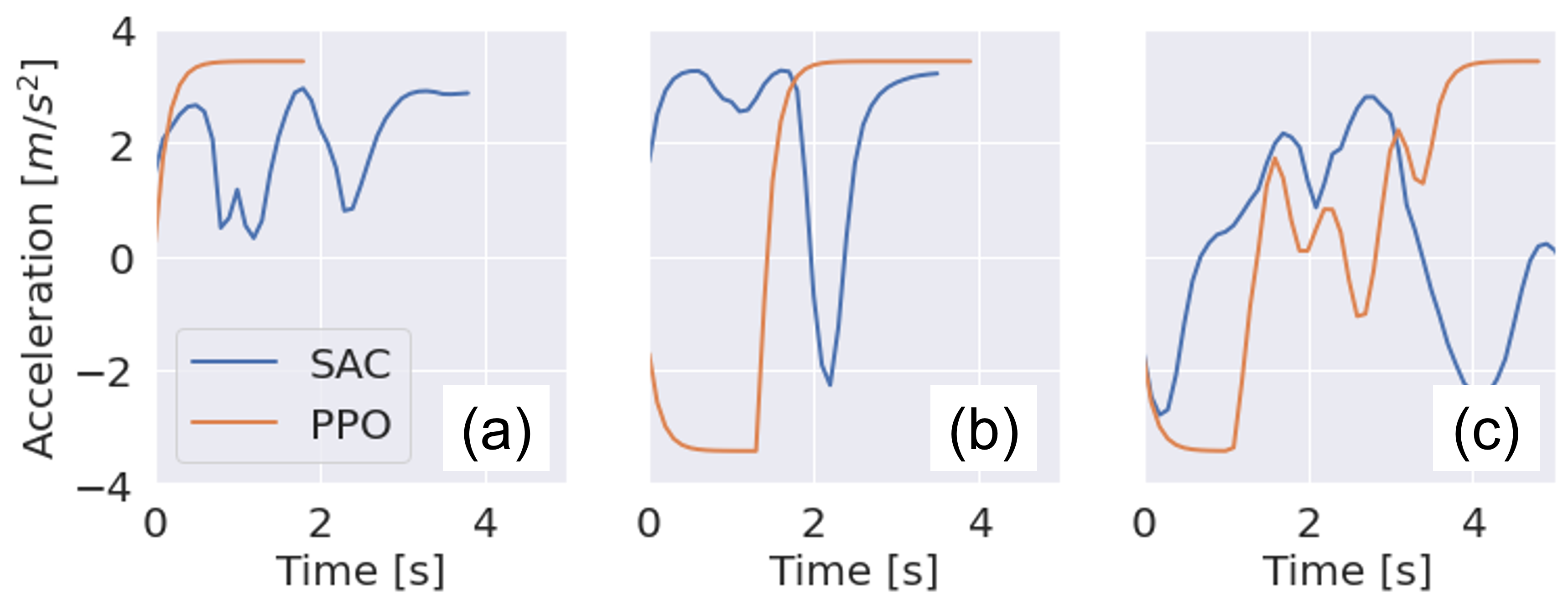}
    \caption{Acceleration profiles for PPO and SVO policies on the same testing episode with SVO values of 0°-(a), 40°-(b), and 80°-(c).}
    \label{fig:accelerations}
    \vspace{-.4 cm}
\end{figure}

In principle, two RL algorithms that solve an MDP problem should both yield optimal policies which achieve the same cumulative reward Fig. \ref{fig:training}. However, actions taken are not necessarily the same. Agents trained with PPO showed smoother acceleration profiles, as shown in Fig. \ref{fig:accelerations}, consistently with DRL theory. For an AV passenger, the policies generated by PPO seem to be more comfortable from an ergonomics perspective, a fact that we intend to investigate in future research. However, SAC has better exploration strategies, rendering it more suitable to solve complex tasks.

\color{black}
Fig. \ref{fig:results} shows the average minimum distance between the ego-vehicle and the pedestrian. The distance increases as the SVO increases, which indicates that the AV has a more altruistic behaviour and yields to the pedestrian. We observed that the policies trained with the SAC algorithm tend to stop much earlier to yield to the pedestrian compared to the PPO algorithm, offering an explanation to why the average minimum distance are significantly larger for such policies.

\color{black}
Overall, the results are consistent with our previous findings \cite{luca21humancentric}, which confirms that the agents are capable of learning behavioural strategies with more complex pedestrian behaviour while still being able to handle risky or unexpected scenarios, which is promising for real world applications. 
\color{black}

\section{Conclusions}
%By Hubert: here is a suggested structure. You can decide if you wish to follow this.

%First paragraph: summarize the whole paper and recall the major contributions.
In this paper we presented an Autonomous Vehicles decision-making strategy for pedestrian collision avoidance based on DRL. We introduced a novel pedestrian model for computer simulation that joins gap-acceptance and social-force models that incorporates a situational awareness risk evaluation to initiate crossing. We demonstrated how our model is capable of handling more complex human models, which is an important prerequisite in order to handle real pedestrians. We have also conducted a comparative analysis of two different model-free DRL algorithms (SAC and PPO) designed for continuous actions spaces applied to our problem.

%Second paragraph: discuss a few points of advantage of the proposed system, and highlight if any of the advantage makes the system better than existing research.
We have shown how PPO policies lead to smoother actions which are more appealing from an ergonomics perspective and offer improvements with respect to previous papers that applied DRL to our problem \cite{li2020deep, deshpande2019deep}. This work also highlights how SVO can be an effective tool to design DRL algorithms in human-machine interaction applications. \color{black}A limitation of our current work is that the SVO policies are trained with discrete SVO values and one would have to switch controllers to alter the ego-vehicle behaviour. We further intend to investigate whether SVO can be used as an input parameter for the neural network, rather than being a fixed parameter at the beginning of each training. This would allow for continuous changes in the car behaviour and the usage of a single controller architecture.

%Third paragraph: discuss a few points of limitations of the proposed method and briefly explain why there are such limitations. e.g. out of our research scope, due to some technical difficulties, etc.
\color{black}
The main assumption within our work is the presence of a single pedestrian. An immediate extension of this work will be to tackle the presence of multiple pedestrians and vehicles. 

%Fourth paragraph: List a number of future directions. For each direction, try to cite papers to say what can be added onto our system or how our system can be adapted for more realistic applications. 
Further, we are looking to improve the state-space-representation and utilise more advanced neural network architectures to validate our model. 
\color{black} In this work, we mainly focused on qualitative results based on simulation to assess the human-likeness of our methods. Our next step is to build up a Virtual Reality environment and perform a subjective human factor analysis with a human-in-the-loop.
\color{black}
\vspace{-0.2cm}
\printbibliography

% your own custom macro containing the $\backslash${\tt{includegraphics}} command to make things
\vspace{-33pt}

\begin{IEEEbiography}[{\includegraphics[width=1in,height=1.25in,clip,keepaspectratio]{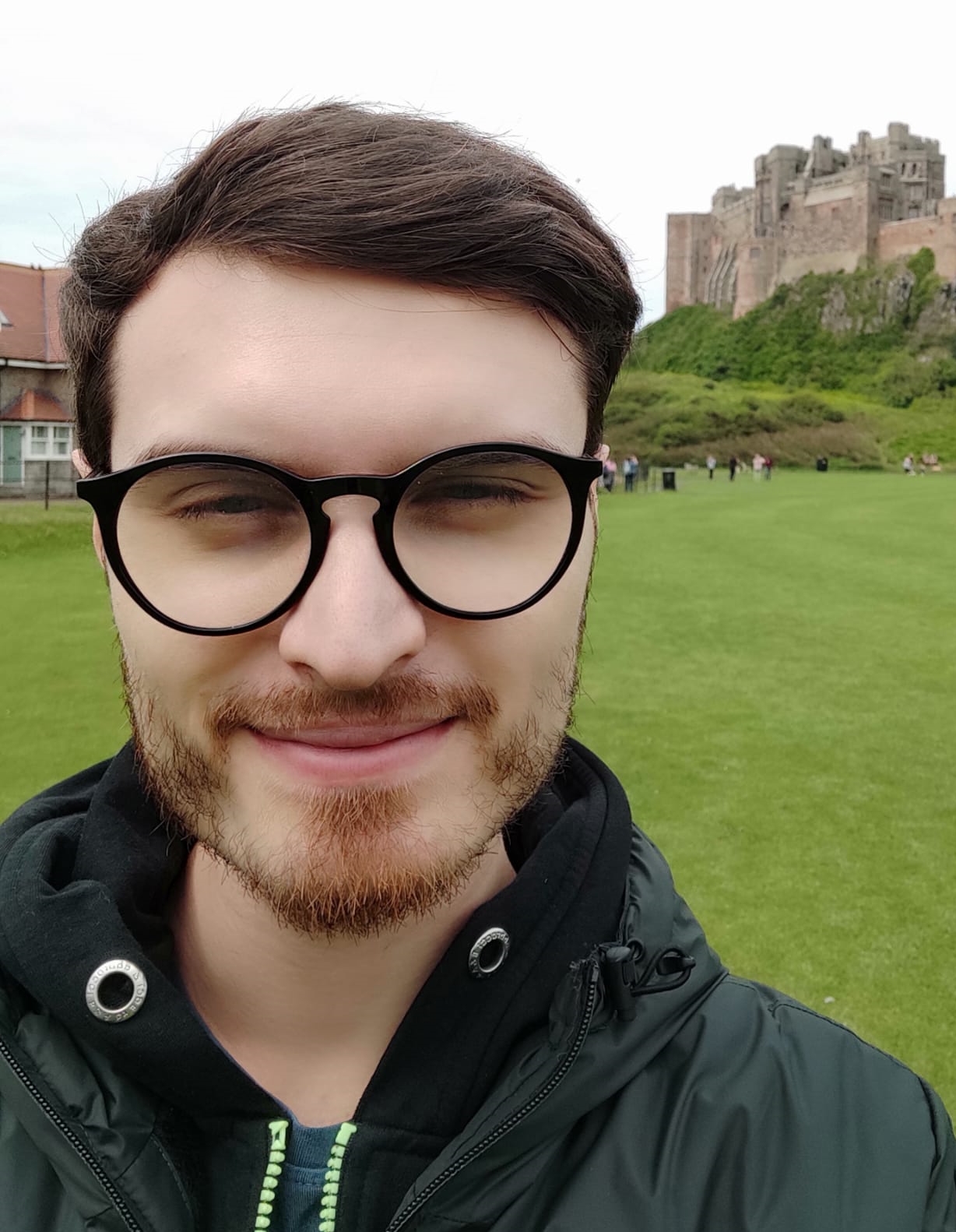}}]{Luca Crosato} received his Bachelor's degree in Mechanical Engineering and the Master’s degree in Robotics and Automation Engineering from the University of Pisa, Italy in 2018. He is currently a 2$^{\text{nd}}$ year PhD student at Northumbria University, UK and Research Assistant at Queen's University Belfast, UK. His research interests include decision-making and control of autonomous vehicles, pedestrian motion simulation and machine learning.
\end{IEEEbiography}
\vspace{-33pt}
\begin{IEEEbiography}[{\includegraphics[width=1in,height=1.25in,clip,keepaspectratio]{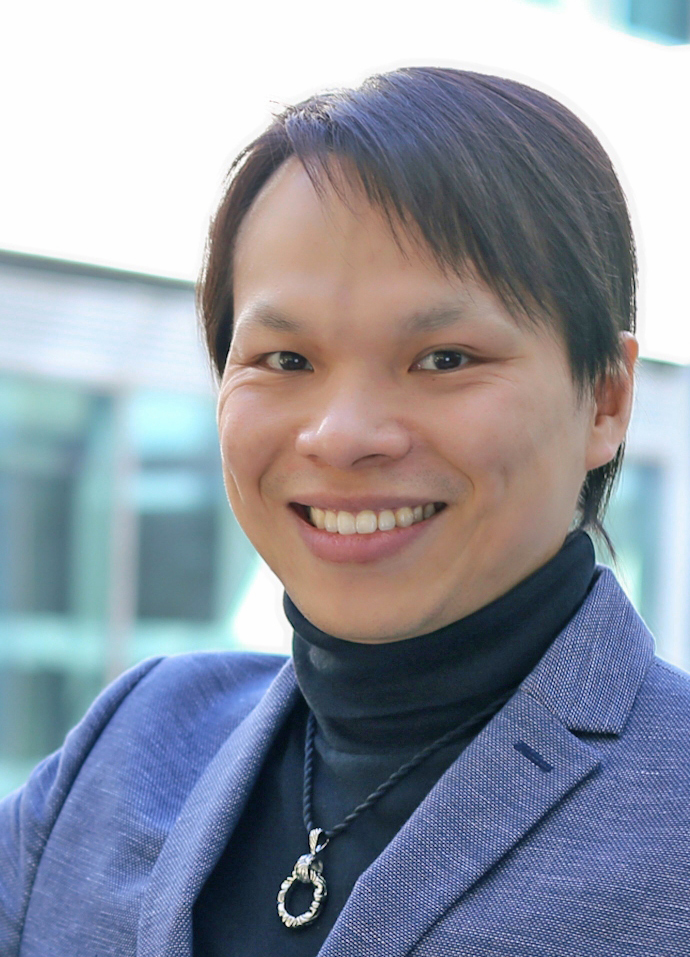}}]{Hubert P. H. Shum} (Senior Member, IEEE) is an Associate Professor in Computer Science at Durham University, UK. Before this, he worked as the Director of Research/Associate Professor/Senior Lecturer at Northumbria University, UK, and a Postdoctoral Researcher at RIKEN, Japan. He received his PhD degree from the University of Edinburgh, UK. His research interests include computer vision, computer graphics, motion analysis and machine learning.
\end{IEEEbiography}

\vspace{-33pt}
\begin{IEEEbiography}[{\includegraphics[width=1in,height=1.25in,clip,keepaspectratio]{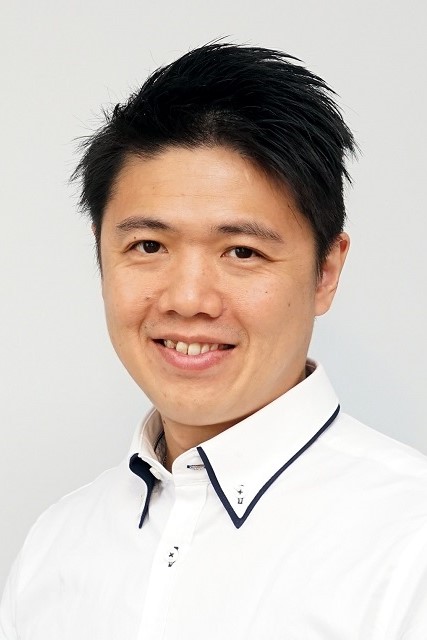}}]{Edmond S. L. Ho} is currently an Associate Professor in the Department of Computer and Information Sciences at Northumbria University, Newcastle, UK. Prior to joining Northumbria University in 2016 as a Senior Lecturer, he was a Research Assistant Professor in the Department of Computer Science at Hong Kong Baptist University. His research interests include Computer Graphics, Computer Vision, Robotics, Motion Analysis, and Machine Learning.
\end{IEEEbiography}
\vspace{-33pt}
\begin{IEEEbiography}[{\includegraphics[width=1in,height=1.25in,clip,keepaspectratio]{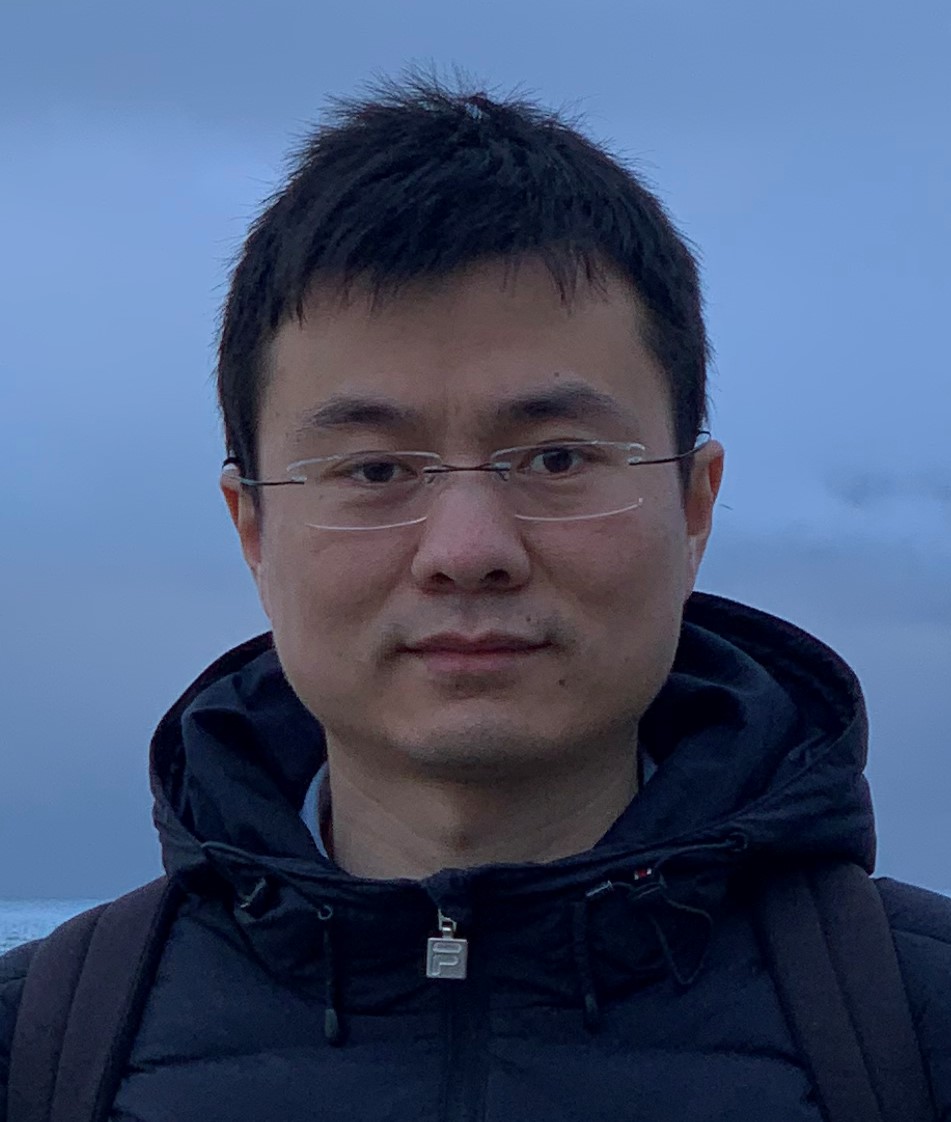}}]{Chongfeng Wei} received the PhD degree in mechanical engineering from the University of Birmingham, UK, in 2015. He is now an assistant professor at Queen's University Belfast, UK. His current research interests include decision making and control of intelligent vehicles, human-centric autonomous driving, cooperative automation, and dynamics and control of mechanical systems. He is also serving as an Associate Editor of IEEE Open Journal of Intelligent Transportation Systems and Associate Editor or IEEE Transactions on Intelligent Vehicles.
\end{IEEEbiography}

\end{document}